\documentclass[10pt,twocolumn,letterpaper]{article}
 \pdfoutput=1
\usepackage{iccv}
\usepackage{times}
\usepackage{epsfig}
\usepackage{graphicx}
\usepackage{amsmath}
\usepackage{amssymb}
\usepackage{algorithm}
\usepackage{algorithmicx}
\usepackage{algpseudocode}
\usepackage{amsmath}
\usepackage[table]{xcolor}
\usepackage{enumitem}
\usepackage{pifont}
\definecolor{lightgray}{gray}{0.9}
\newcommand{\cmark}{\ding{51}}%
\usepackage[pagebackref=true,breaklinks=true,letterpaper=true,colorlinks,bookmarks=false]{hyperref}

\usepackage[breaklinks=true,bookmarks=false]{hyperref}

\iccvfinalcopy 

\begin{document}

\title{Self-Supervised Video Representation Learning with Meta-Contrastive Network}

\author{Yuanze Lin$^1$\thanks{The work was done when the author was with MSRA as an intern.} \quad \quad Xun Guo$^2$ \quad \quad Yan Lu$^2$\\
$^1$University of Washington, \quad $^2$Microsoft Research Asia\\
{\tt\small yuanze@uw.edu, \quad \{xunguo, yanlu\}@microsoft.com}\\
}

\maketitle

\begin{abstract}
Self-supervised learning has been successfully applied to pre-train video representations, which aims at efficient adaptation from pre-training domain to downstream tasks. Existing approaches merely leverage contrastive loss to learn instance-level discrimination. However, lack of category information will lead to hard-positive problem that constrains the generalization ability of this kind of methods. We find that the multi-task process of meta learning can provide a solution to this problem. In this paper, we propose a \textbf{M}eta-\textbf{C}ontrastive \textbf{N}etwork (MCN), which combines the contrastive learning and meta learning, to enhance the learning ability of existing self-supervised approaches. Our method contains two training stages based on model-agnostic meta learning (MAML), each of which consists of a contrastive branch and a meta branch. Extensive evaluations demonstrate the effectiveness of our method. For two downstream tasks, \textit{i.e.}, video action recognition and video retrieval, MCN outperforms state-of-the-art approaches on UCF101 and HMDB51 datasets. To be more specific, with R(2+1)D backbone, MCN achieves Top-1 accuracies of \textbf{84.8\%} and \textbf{54.5\%} for video action recognition, as well as \textbf{52.5\%} and \textbf{23.7\%} for video retrieval.

\end{abstract}
\section{Introduction}
Convolutional Neural Networks (CNNs) have brought unprecedented success for supervised video representation learning \cite{carreira2017quo, feichtenhofer2019slowfast, feichtenhofer2016convolutional, wang2016temporal, lin2019tsm} . However, labeling large-scale video data requires huge human annotations, which is expensive and laborious. How to learn effective video representations by leveraging unlabeled videos is an important yet challenging problem. The recent progress of self-supervised learning for image provides an efficient solution to this problem \cite{he2020momentum, tian2019contrastive, henaff2020data, chen2020simple}, which proposed to use contrastive loss \cite{gutmann2010noise, hadsell2006dimensionality, wang2015unsupervised, wu2018unsupervised, hjelm2018learning} to discriminate different data samples. 

\begin{figure}[H]
\centering
\includegraphics[scale=0.345]{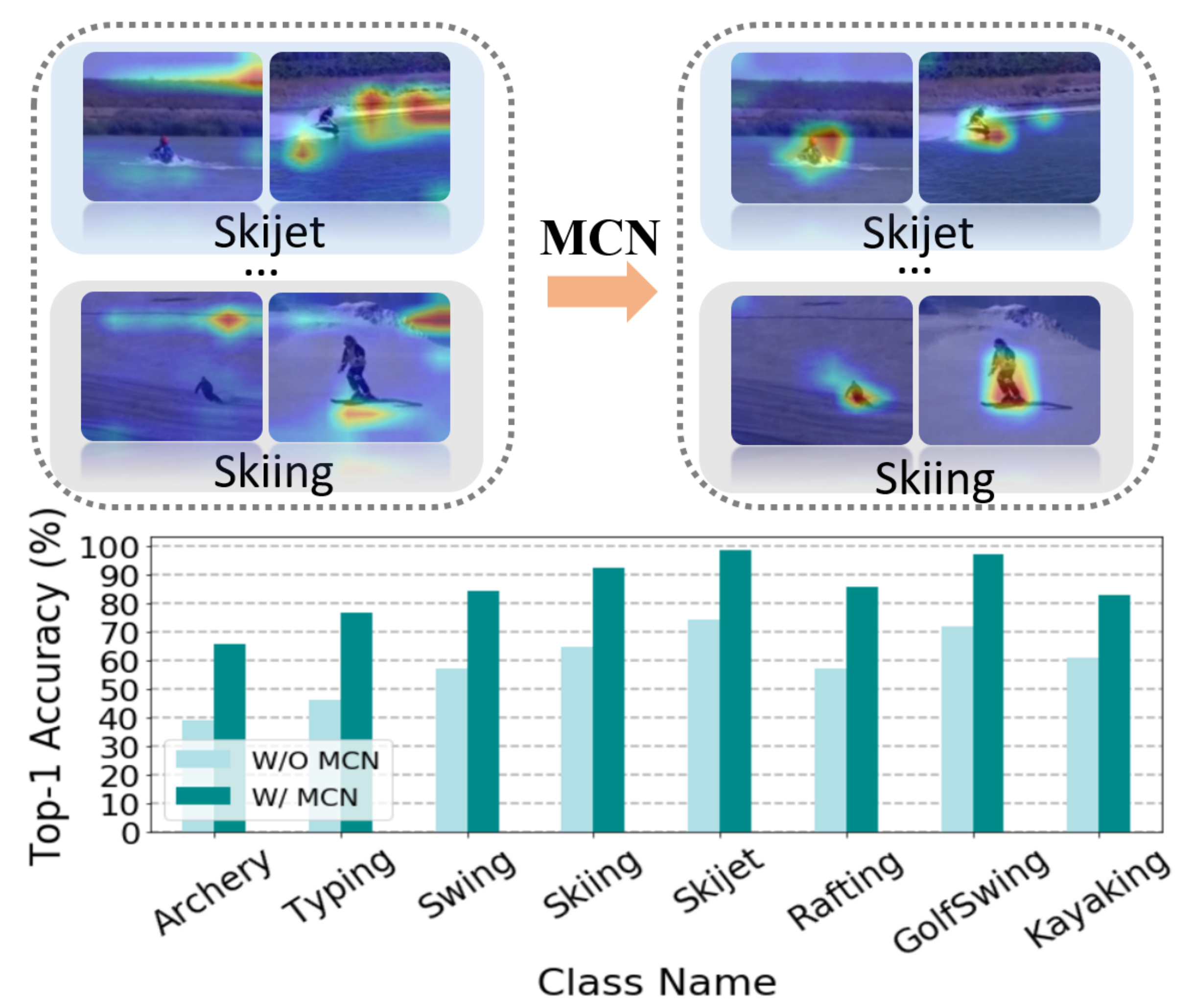}
\caption{\textbf{Comparison between models trained without and with MCN on UCF101 \cite{soomro2012ucf101}}. The top row shows the activation maps produced by conv5 layer of R(2+1)D backbone using the method of \cite{zagoruyko2016paying}. By using our proposed MCN, the learned representations can  capture motion areas more accurately. The bottom row shows top-1 accuracies of models trained without and with MCN approach.}
\label{fig:sample}
\centering
\end{figure}

 This instance-based contrastive learning has also been applied to videos as pre-training, and achieved excellent performance on downstream tasks such as video action recognition and video retrieval \cite{Han20, tao2020self, miech2020end}. However, it has the inherent limit of lacking the common category information. The instance-based discrimination process takes each video sample as an independent class, so that distance between two video samples will be pushed away by contrastive loss even if they belong to the same category. This drawback reduces the generalization of the pre-training parameters. Consequently, the efficiency of the supervised fine-tuning for the downstream tasks will also be damaged. How to improve the generalization of contrastive self-supervised learning and make the learned parameters easily adapt from pre-training domain to fine-tuning domain for various new tasks is still challenging.  

Meta learning has demonstrated the capability of fast adaptation on new tasks with only a few training samples. The characteristic of meta learning, specifically model-agnostic meta learning (MAML) \cite{finn2017model}, might help contrastive self-supervised video learning in two aspects. Firstly, instance-based discrimination takes each video as a class, so that it is convenient to create numerous sub-tasks for meta learning to improve the model generalization. Secondly, the goal of meta learning is “learn to learn”, which means that it provides good initialization for fast adaptation on a new task. This perfectly meets the requirements of contrastive video representation learning, which is taken as a pre-training method. Therefore, combining meta learning and self-supervised learning might benefit to video representation learning.

In this paper, we propose a novel \textbf{M}eta-\textbf{C}ontrastive \textbf{N}etwork (MCN), which leverages meta-learning to improve the generalization and adaptation ability of contrastive self-supervised video learning on downstream tasks. The proposed MCN contains two branches, \textit{i.e.}, contrastive branch and meta branch, which establishes a multi-task learning process to enhance the instance discrimination. Meanwhile, we design a two-stage training process based on MAML to improve the learning capability of MCN. Our method outperforms state-of-the-art methods and achieves significant performance boost. 

The main contributions of this paper are summarized as follows.
\begin{enumerate}[label=\arabic*)]
\item We propose a novel \textbf{M}eta-\textbf{C}ontrastive \textbf{N}etwork (MCN), which can significantly improve the generalization of the video representations learned in self-supervised learning manner.

\item We fully investigate the benefits of combining meta learning with self-supervised video representation learning and conduct extensive experiments to make proposed approach better understood.

\item We evaluate our method on mainstream benchmarks for action recognition and retrieval tasks, which demonstrate that our proposed method can achieve state-of-the-art or comparable performance with other self-supervised learning approaches.
\end{enumerate}

\section{Related Work}
\subsection{Pretext Tasks} 

Early self-supervised learning approaches mainly focus on designing handcrafted pretext tasks for images, such as predicting the rotation of transformed images \cite{gidaris2018unsupervised}, image jigsaw \cite{doersch2015unsupervised}, count of learned features \cite{noroozi2017representation}, image colorization \cite{zhang2016colorful}, relative positions \cite{doersch2015unsupervised} and so on. 

After that, many self-supervised learning approaches about video data flourish. Due to the extra temporal dimension of video data, there are many pretext tasks specifically designed for temporal prediction, such as frame rate prediction \cite{wang2020self},  pace prediction \cite{benaim2020speednet},  frame ordering prediction \cite{xu2019self, lee2017unsupervised, fernando2017self} and motion statistics prediction \cite{wang2019self}. 

These pretext tasks make models achieve better discriminative ability, which is important for downstream tasks.

\subsection{Contrastive Self-Supervised Learning}
Contrastive self-supervised learning has been proved great potential in unlabeled data \cite{he2020momentum, tian2019contrastive, henaff2020data, chen2020simple, misra2020self, han2019video}. Thanks to contrastive self-supervised learning approaches, model can be empowered to distinguish samples from different domains without labels.

There are some prior works in this area. He \textit{et al.} \cite{he2020momentum} proposed a momentum dictionary to store and pop out learned features on the fly for images, so that the number of stored features can be extremely expanded. Chen \textit{et al.} \cite{chen2020simple} proposed a simplified constrastive self-supervised image learning framework including only major components that benefit the learned representations. Tian \textit{et al.} \cite{tian2019contrastive} presented a contrastive multi-view coding (CMC) approach for video representation learning, which uses different views of input videos to maximize the instance-level distinction. Our approach adopts CMC with two views, \textit{i.e.}, RGB view and residual view, as baseline.

\subsection{Meta Learning}
Numerous research about meta learning has been presented for few-shot tasks. Finn \textit{et al.} \cite{finn2017model} proposed an important meta learning method called model-agnostic meta learning (MAML), which can be combined with any learning approaches trained with gradient descent. Some variants of MAML, \textit{e.g.,} Reptile \cite{nichol2018first} and iMAML \cite{rajeswaran2019meta}, can not only significantly save the training time, but also achieve comparable performance with MAML.

Recently, researchers start to focus on applying meta learning approaches to computer vision tasks, such as object tracking and face recognition \cite{guo2020learning, wang2020tracking, wang2019meta}. In this paper, we utilize MAML to improve the performance of contrastive self-supervised video learning. Different from prior efforts, we try to enhance the adaptation between self-supervised pre-training domain and supervised fine-tuning domain, which is more challenging.  

\begin{figure*}
\begin{center}
\includegraphics[height=92mm,width=175mm]{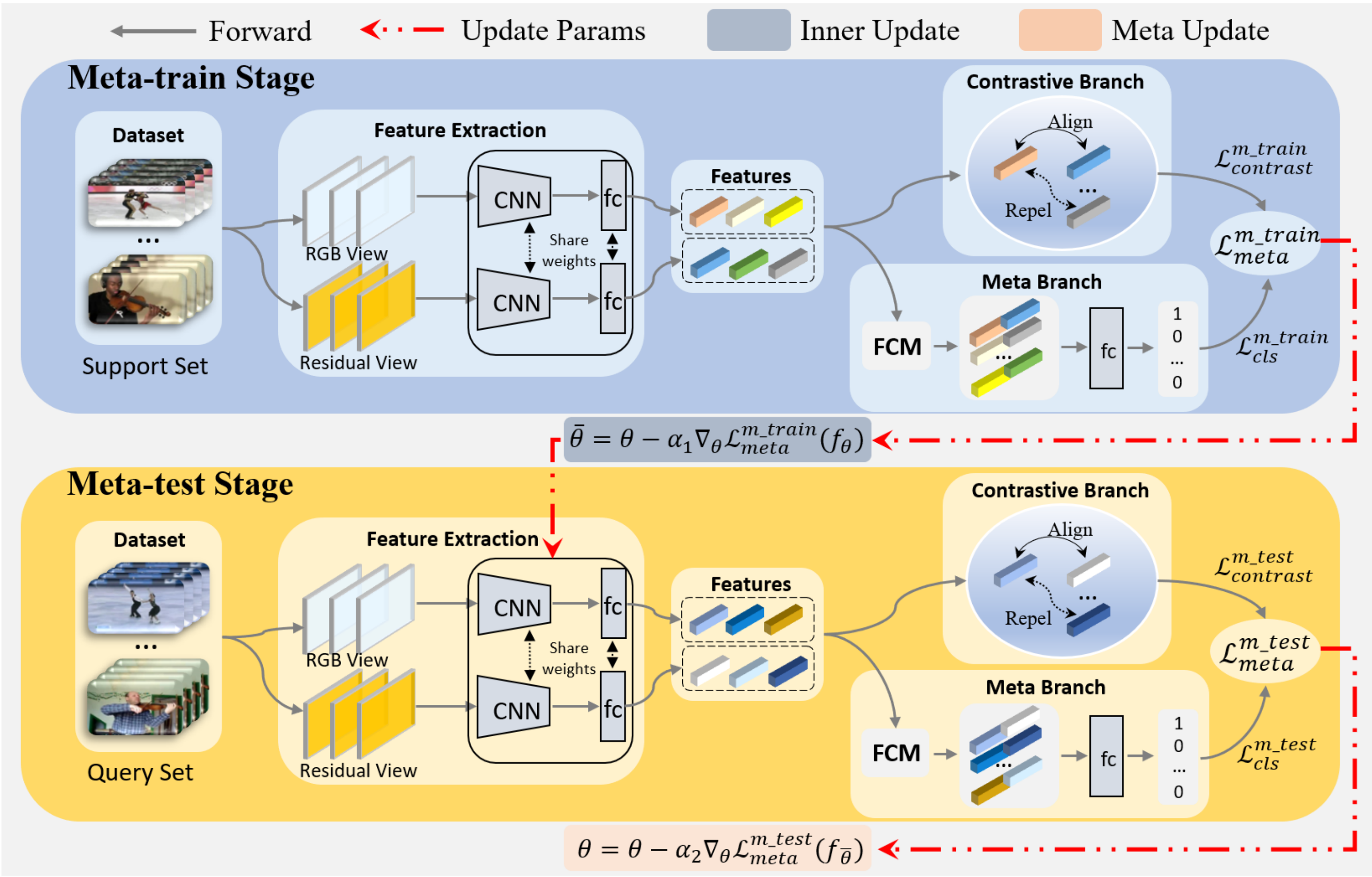}
\end{center}
   \caption{\textbf{The illustration of Meta-Contrastive Network}. For simplicity, only 3 input videos are used for illustration. There are two stages of MCN, including meta-train and meta-test stages. Model is parameterized by $\theta$ initially. $\alpha_{1}$ and $\alpha_{2}$ represent learning rates. Note that fully connected (fc) layer in meta branch is different from that in feature extraction module. FCM is feature combination module, which generates binary classification features for meta branch.} 
\label{fig:framwork}
\end{figure*}

\section{Meta-Contrastive Network}
In this section, we describe our proposed meta-contrastive network (MCN) in details. In section \ref{framwork}, we introduce the framework of MCN and the two-stage training process. In section \ref{main_branch}, we elaborate the contrastive branch of MCN. In section \ref{meta_branch}, the details of meta branch are clarified. In section \ref{loss_and_optimization}, we introduce the losses and optimizations in MCN. Finally, In section \ref{implementation}, implementation details are explained.

\subsection{Framework}
\label{framwork}
We build our framework by employing the contrastive multi-view coding (CMC) \cite{tian2019contrastive} as baseline. Multi-view input is proved to be efficient for instance-based video contrastive learning since different views, \textit{i.e.}, transformations, from the same video can increase positive samples and make contrastive learning more efficient. We adopt two views in our framework, \textit{i.e.}, RGB view and residual view, which have been proved to be extremely efficient views \cite{tao2020rethinking}. There are two branches in our framework as shown in Figure \ref{fig:framwork}, \textit{i.e.}, contrastive branch and meta branch. The contrastive branch performs contrastive learning, and the meta branch performs a couple of binary classifications for efficient meta learning. A binary classification is very similar with a pretext task that predicts whether the input two features come from the same video sample. 

We employ a two-stage training process including meta-train and meta-test. Training data is split into train set, \textit{i.e.}, \textit{support set}, and test set, \textit{i.e.}, \textit{query set}. In meta-train stage, the videos from support set are used for inner update, in which the updated parameters will be used in meta-test stage for feature extraction. In meta-test stage, the videos from query set are used with the inner updated parameters for meta-update, which updates the initial parameters of meta-train stage for the next training iteration of MCN. 

\subsection{Contrastive Branch}
\label{main_branch}
The contrastive branch constructs a feature bank for positive and negative samples by collecting the extracted features for both of the two views, and calculates contrastive loss, \textit{i.e.}, NCE loss \cite{gutmann2010noise}. The RGB view contains the sampled RGB video frames from a video clip, and the residual view contains the differences between two consecutive RGB frames. A residual frame is calculated as:

	\begin{align}
	\label{eq:res_img}
	{Frame_{n}^{Res}} = |{Frame_{n}^{RGB}} -        
	{Frame_{n+1}^{RGB}}|,
	\end{align}

\noindent where $Frame_{n}^{Res}$ represents residual frame; $Frame_{n}^{RGB}$ represents RGB frame; $n$ is the index of the sampled frame.

The reason why residual view is efficient may be that it can reflect the motions of the video clip to some extent and provides complementary information to RGB view. For example, when there are two different video clips with the same action, they may have similar residual view. This will implicitly increase the hard positive in contrastive learning process.    

\subsection{Meta Branch}
\label{meta_branch}
Constrative learning suffers from hard-positive problem and hard-negative problem, which are even worse in self-supervised video learning. For example, there may exist videos that contain totally different scenes and objects but the same actions and events. There also exists videos that contain similar scenes and objects but different actions and events. Theoretically, meta learning can alleviate this problem due to the multi-task learning process. For this purpose, we design the meta branch consisting of a feature combination module (FCM) and several binary classification tasks, which can predict whether a feature pair belongs to the same video clip.  

As shown in Figure \ref{fig:framwork}, by concatenating two features of input video samples in FCM, several instance/binary classification tasks can be constructed in meta branch. The corresponding labels can be easily created for training. Figure \ref{fig:sample} shows an example of creating classification task with FCM for two video samples $v_{1}$ and $v_{2}$. If one concatenated feature is from the different views of the same video, the label will be true, otherwise the label will be false.

We design the binary classification lies in two reasons. Firstly, the binary classification loss can be complementary with contrastive loss to better learn the instance discrimination. Secondly, the binary classification enables efficient combination of contrastive learning branch and meta learning branch to improve the generalization through multi-task learning process. 

\begin{figure}[H]
\centering
\includegraphics[scale=0.28]{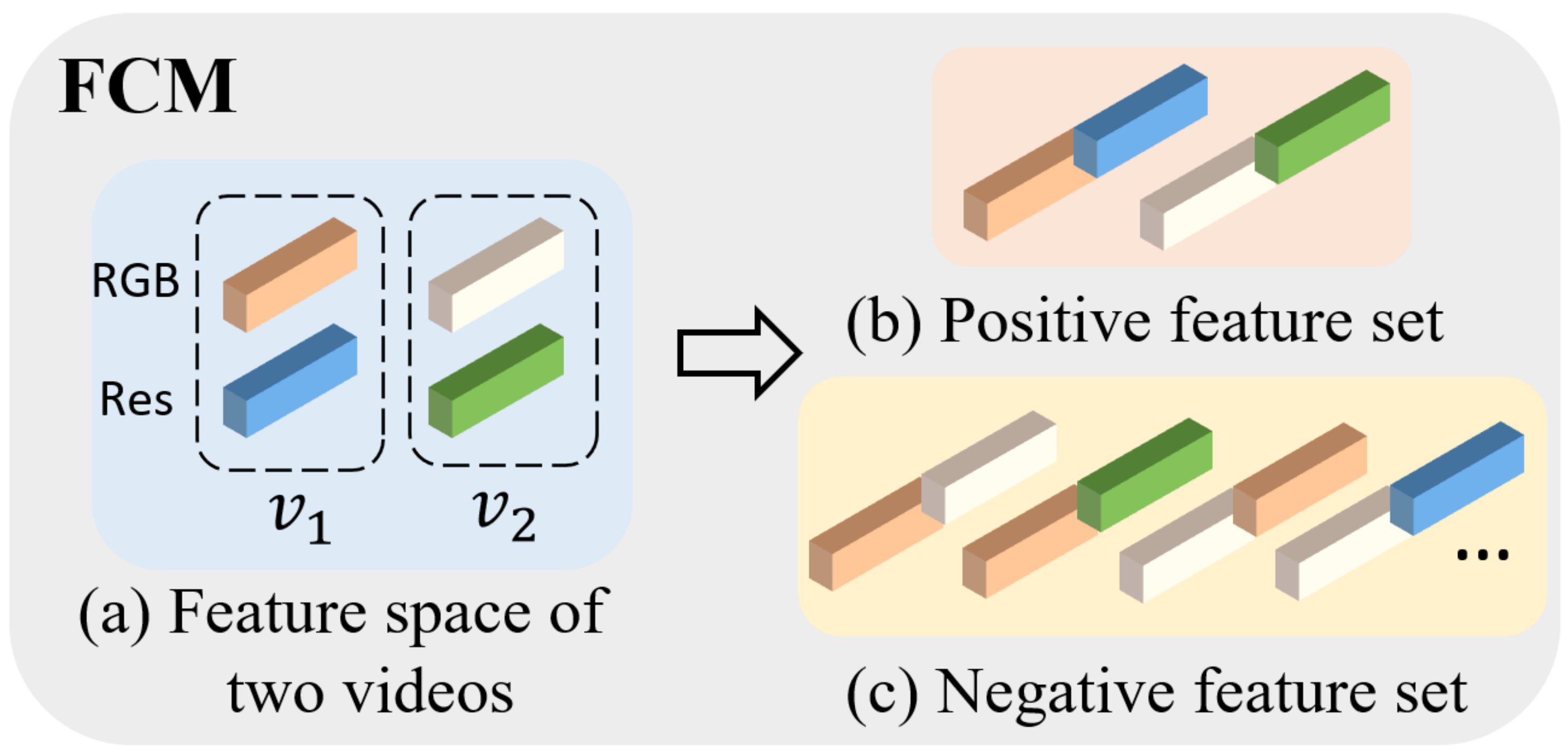}
\caption{\textbf{An example of FCM}. $v_{1}$ and $v_{2}$ are two video samples. (a) is feature space of the two videos. RGB and Res mean features extracted from RGB view and Residual view respectively. (b) is positive feature set, whose label is true. (c) is negative feature set, whose label is false.}
\label{fig:sample}
\centering
\end{figure}
 
\subsection{Meta Loss and Optimization}
In order to ease the training process of MCN method, we incorporate both metric losses, \textit{i.e.}, contrastive loss and cross-entropy loss, \textit{i.e.}, classification loss, and propose the combined meta loss for final optimization.

\label{loss_and_optimization}
\textbf{Contrastive Loss.} Contrastive learning aims to separate features from different samples. In our method, we employ the contrastive loss form from CMC \cite{tian2019contrastive} as the objective of the contrastive branch. Specifically, two different views of the same sample, \textit{e.g.}, $\{$$x_{i}^{1}, x_{i}^{2}$$\}$, are treated as positive, while the views from different samples, \textit{e.g.}, $\{$$x_{i}^{1}, x_{j}^{2}$$\}$ $(i$$\not=$$j)$, are regarded as negative. 

A value function $h_{\theta}$ is used so that positive pairs  have high score, and negative pairs obtain low score. To be more concrete, after feature $z_{i}^{1}$ is extracted by the model, function $h_{\theta}$ is trained on a feature set $Z$ $=$ $\{$$z_{1}^{2}$, $z_{i}^{2}$, ..., $z_{k+1}^{2}$$\}$, which consists of one positive sample $z_{i}^{2}$ and $k$ negative samples, so that the positive sample can be easily picked out from $Z$. The contrastive loss can be formulated as:
	\begin{align}
	\label{eq:contrastive_loss}
	{L_{contrast}} = {-log{\frac{h_{\theta}(\{{z_{i}^{1}, z_{i}^{2}}\})}
	{\sum_{j=1}^{k+1}h_{\theta}(\{{z_{i}^{1}, z_{j}^{2}}\})}}},
	\end{align}

\noindent where $L_{contrast}$ denotes contrast loss for contrastive branch, and $k$ is the number of negative samples. $z_{i}^{1}$ and $z_{i}^{2}$ mean extracted features of two different views from $i$th sample. $h_{\theta}(\cdot)$ can be formulated as:
	
	\begin{align}
	\label{eq:f}
	h_{\theta}(\{z_{i}^{1}, z_{j}^{2}\}) = exp(\frac{z_{i}^{1}\cdot{z_{j}^{2}}}
	{\left\|z_{i}^{1}\right\|\cdot{\left\|z_{j}^{2}\right\|}\cdot{\tau}}),
	\end{align}

\noindent where $h_{\theta}(\cdot)$ is cosine similarity of two features and $\tau$ is the parameter for dynamically controlling the range.

\textbf{Classification Loss.}
Meta branch of MCN performs instance/binary classification. We use binary cross-entropy loss (BCE) as our classification loss, which can be formulated as:

	\begin{align}
	\label{eq:cls}
	{L_{cls}} = {\sum_{i=1}^{N}-y_{i}log{\hat{y}}} - {(1-y_{i})log(1-\hat{y_{i}})},
	\end{align}

\noindent where $N$ is the number of concatenated features. In our approach, FCN concatenates features from 4 video clips in each batch and each video has two features. $y_{i}$ is the label of $i$th concatenated feature. $\hat{y}_{i}$ is the output of the fully connected layer of meta branch. 

\textbf{Meta Loss.} Contrastive loss and classification loss from the two branches are combined together to get the final meta loss, which is defined as:
	\begin{align}
	\label{eq:meta_loss}
	{L_{meta}} = \alpha \cdot {L_{cls}} + (1-\alpha) \cdot {L_{contrast}},
	\end{align}

\noindent where $\alpha$ is a hyper-parameter used to control the relative impact of binary classification loss $L_{cls}$ and contrast loss $L_{contrast}$ respectively. The meta loss is used to update weights in meta-train and meta-test stages.

\textbf{Optimization.} During meta-train or meta-test stage, meta loss $L_{meta}$ is used to optimize model parameters $\theta$ with gradient descent. In meta-train stage, $L_{meta}$ achieved from support set is used to get updated parameters $\bar{\theta}$, which is denoted as inner update. And in meta-test stage, $L_{meta}$ obtained from query set is used to update $\theta$, which is denoted as meta update. Step sizes of gradient descent for the two optimization stages are the same as learning rate.
    
\subsection{Implementation Details}
\label{implementation}
We will further explain implementation details of MCN in this section. In specific, our proposed MCN consists of two stages, \textit{i.e.}, meta-train and meta-test stages. The whole process of MCN is explained as follows.

\textbf{Initialize.} A pre-trained model ${f(\theta)}$ parametrized by $\theta$, dataset D, support set $D_{s}$, query set $D_{q}$, batch size $B$, $D$ = $D_{s}$ $\cup$ $D_{q}$. 

\textbf{Input.} Sample $B$ input videos $X_{sup}$ from support set $D_{s}$. Sample $B$ input videos $X_{que}$ from query set $D_{q}$.

\textbf{Meta-train.} Feed sampled videos $X_{sup}$ into network ${f(\theta)}$ to extract features. Use these features to compute contrastive loss $L_{contrast}^{m{\_}train}$ by Equation \ref{eq:contrastive_loss}. Use FCM to combine these features and get a new feature set $S_{m{\_}train}$. Set $S_{m{\_}train}$ is used to compute classification loss $L_{cls}^{m{\_}train}$ by Equation \ref{eq:cls}, then $L_{cls}^{m{\_}train}$ and $L_{contrast}^{m{\_}train}$ are used to compute $L_{meta}^{m{\_}train}$ by Equation \ref{eq:meta_loss}.

\textbf{Inner Update.} Use $L_{meta}^{m{\_}train}$ to update model parameters by gradient descent. The updated process is $\bar{\theta}$ $=$ $\theta - \alpha_{1} {\nabla}_{\theta}L_{meta}^{m{\_}train}(f_{\theta})$, where $\alpha_{1}$ is the same as learning rate.

\textbf{Meta-test.} Feed sampled videos $X_{que}$ into network ${f(\bar{\theta})}$ to extract features. Use these features to compute contrastive loss $L_{contrast}^{m{\_}test}$ by Equation \ref{eq:contrastive_loss}. Use FCM to combine these features and get a new feature set $S_{m{\_}test}$. Set $S_{m{\_}test}$ is used to compute classification loss $L_{cls}^{m{\_}test}$ by Equation \ref{eq:cls}, then $L_{cls}^{m{\_}test}$ and $L_{contrast}^{m{\_}test}$ are used to compute $L_{meta}^{m{\_}test}$ by Equation \ref{eq:meta_loss}.

\textbf{Meta Update.} Use $L_{meta}^{m{\_}test}$ to update the final model parameters by gradient descent. Corresponding updated process is ${\theta}$ $=$ $\theta - \alpha_{2} {\nabla}_{\theta}L_{meta}^{m{\_}test}(f_{\bar{\theta}})$, where $\alpha_{2}$ is the same as learning rate. 

\section{Experiments}
\label{experiments}
\subsection{Datasets}
We evaluate our approach on three video classification datasets including UCF101 \cite{soomro2012ucf101}, HMDB51 \cite{kuehne2011hmdb} and Kinetics-400 \cite{kay2017kinetics}.

\textbf{UCF101} \cite{soomro2012ucf101} is a dataset that has 101 action categories, containing 13320 videos totally. There are 3 splits on this data set \cite{benaim2020speednet, xu2019self}. In our experiments, we use train split 1 as self-supervised pre-training dataset, and train/test split 1 for fine-tuning/evaluation.  

\textbf{HMDB51} \cite{kuehne2011hmdb} has around 7000 videos with 51 video action classes, which is relatively small compared to UCF101 and Kinetics \cite{kay2017kinetics}. It also has 3 splits. We use split 1 for fine-tuning and evaluation.

\textbf{Kinetics-400} \cite{kay2017kinetics} is a popular benchmark for action recognition collected from Youtube, which contains 400 action categories. There are totally 300K video samples, which  are  divided  to  240K,  20K  and  40K for training, validation and test sets respectively. In our paper, we only use train split as our pre-training dataset

\subsection{Experimental Setup}
\textbf{Data Pre-processing.} We randomly sample 32 continuous frames from each video as the input of MCN. If the original videos are not long enough, the first frame will be repeated. The sampled original frames are treated as RGB view, and the residual frames generated by Equation \ref{eq:res_img} is treated as residual view. The original frames will be randomly cropped and resized into 128$\times$128. Meanwhile, Gaussian blur, horizontal flips and color jittering are also used for augmentation.

\textbf{Backbones.} Three main-stream network structures, \textit{i.e.}, S3D \cite{xie2018rethinking}, R3D-18 \cite{hara2018can, tran2018closer} and R(2+1)D \cite{tran2018closer} are used as the backbones of MCN in ablation experiments. For video action recognition and video retrieval tasks, only the results of R3D-18 and R(2+1)D are reported.

\textbf{Self-Supervised Learning.} We train our models using 4 NVIDIA Tesla P40 around 500 epochs. Initial learning rate is 0.01 and weight decay is 0.001. $\alpha$ is set to 0.2. We use the batch sizes of 28 and 80 for R(2+1)D and R3D-18 respectively.

\textbf{Fine-tuning.} After finishing self-supervised learning stage, we fine-tune the pre-trained models on UCF101 or HMDB51 around 300 epochs. A new fully connected layer will be added to the end of the pre-trained backbone for classification. Learning rate is set as 0.02. And the batch sizes are 72 and 200 for R(2+1)D and R3D-18 respectively.

\textbf{Evaluations.} Evaluations of proposed method are conducted on video action recognition and video retrieval tasks. For video action recognition, the top-1 accuracies on UCF101 \cite{soomro2012ucf101} and HMDB51 \cite{kuehne2011hmdb} are reported. In order to further validate our proposed MCN, we also show the results of linear probe results, in which the weights of self-supervised learning model are fixed and only the fully-connected layers for supervised classification are fine-tuned. For video retrieval, top-1, top-5, top-10, top-20 and top-50 accuracies are compared with existing approachs.

\subsection{Ablation Studies}
To fully investigate and understand the concept of MCN, we conduct ablation experiments to demonstrate how each design of MCN affects the overall performance.
	
\textbf{Comparison with Baseline.} We compare the action recognition results of self-supervised training with and without MCN in Table \ref{tab:ablation_ucf101}. We use three backbones, \textit{i.e.}, S3D, R(2+1)D and R3D-18, to demonstrate the performance boost of MCN. As shown in this table, the accuracies of baseline with S3D, R(2+1)D and R3D-18 are 76.7\%, 77.3\% and 78.6\% on UCF101 dataset respectively. while the accuracies of MCN are \textbf{82.9\%}, \textbf{84.8\%} and \textbf{85.4\%} respectively. Relevant results of HMDB51 dataset can also be observed. There is consistent performance boost when using MCN on different backbones and data sets.

Furthermore, we also evaluate the linear probe results in Table \ref{tab:ablation_probe}, in which only the fully connected layers are fine-tuned. Significant performance boost of MCN can also be observed on both UCF101 and HMDB51 dataset. 

	\begin{table}[htbp]
		\centering
		\resizebox{1.0\linewidth}{!}{%
			\begin{tabular}{|l|c|c|c|}
				\hline 
				\textbf{Methods} & \textbf{Backbone}  & \textbf{UCF101(\%)}  & \textbf{HMDB51(\%)}  \\ 
				\hline
				\hline
				Ours (Baseline) & S3D & 76.7  & 45.5 \\ %
				\cellcolor{gray!25}Ours (+ MCN) & \cellcolor{gray!25}S3D & \cellcolor{gray!25}\textbf{82.9} & \cellcolor{gray!25}\textbf{53.8} \\ %
				\hline
				Ours (Baseline) & R(2+1)D & 77.3  & 46.2 \\ %
				\cellcolor{gray!25}Ours (+ MCN) & \cellcolor{gray!25}R(2+1)D & \cellcolor{gray!25}\textbf{84.8}  & \cellcolor{gray!25}\textbf{54.5} \\ %
				\hline
				Ours (Baseline) & R3D-18 & 78.6  & 47.1 \\ %
				\cellcolor{gray!25}Ours (+ MCN) & \cellcolor{gray!25}R3D-18 & \cellcolor{gray!25}\textbf{85.4}  & \cellcolor{gray!25}\textbf{54.8} \\ %
				\hline
			\end{tabular}
		}
		\vspace{0.05cm}
		\caption{Comparisons between MCN and baseline with different backbones on video action recognition task.} 
		\label{tab:ablation_ucf101}
	\end{table}

	\begin{table}[htbp]
		\centering
		\resizebox{1.0\linewidth}{!}{%
			\begin{tabular}{|l|c|c|c|}
				\hline 
				\textbf{Methods}  & \textbf{Backbone} & \textbf{UCF101(\%)}   & \textbf{HMDB51(\%)}  \\
				\hline
				\hline
				Ours (Baseline)  & S3D   & 62.4 & 33.5 \\ %
				\cellcolor{gray!25}Ours (+MCN)  & \cellcolor{gray!25}S3D   & \cellcolor{gray!25}\textbf{71.6} & \cellcolor{gray!25}\textbf{40.8} \\ %
				\hline
				Ours (Baseline)  & R(2+1)D   &  64.2  & 35.6\\ %
				\cellcolor{gray!25}Ours (+MCN)  & \cellcolor{gray!25}R(2+1)D   & \cellcolor{gray!25}\textbf{72.4}  & \cellcolor{gray!25}\textbf{41.2}\\ %
				\hline
				Ours (Baseline) & R3D-18   & 64.6 & 37.3 \\ %
				\cellcolor{gray!25}Ours (+MCN)  & \cellcolor{gray!25}R3D-18   & \cellcolor{gray!25}\textbf{73.1} & \cellcolor{gray!25}\textbf{42.9} \\ %
				\hline
			\end{tabular}
		}
		\vspace{0.05cm}
		\caption{Linear probe evaluation results of different backbones on video action recognition task.} 
		\label{tab:ablation_probe}
	\end{table}
	
\textbf{Influence of $\alpha$.} As depicted in Equation \ref{eq:meta_loss}, $\alpha$ is introduced to modulate meta loss. We also conducted experiments to demonstrate the influence of this hyper-parameter. Table \ref{tab:hyper} shows the results of 4 settings of $\alpha$ with R(2+1)D backbone.

	\begin{table}[htbp]
		\centering
		\resizebox{0.45\linewidth}{!}{%
			\begin{tabular}{|l|c|}
				\hline 
			  	\textbf{  Settings} & \textbf{UCF101(\%)}  \\ 
				\hline
				\hline
				$\alpha=0.1$  &  84.1 \\ %
				\cellcolor{gray!25}$\alpha=0.2$  &\cellcolor{gray!25}\textbf{84.8}  \\ %
				$\alpha=0.3$  &  83.4 \\ %
				$\alpha=0.4$  &  82.7  \\ %
				\hline
			\end{tabular}
		}
		\vspace{0.25cm}
		\caption{Results of different $\alpha$ settings on UCF101 dataset.} 
		\label{tab:hyper}
	\end{table}
We can observe that setting $\alpha$ as 0.2 shows the best performance. Therefore, we set $\alpha$ as 0.2 in all our experiments.

\textbf{Influence of Input Frames.} For self-supervised video representation learning, the number of input frames for each video clip may affect the final performance. Therefore, we tested different numbers for quantitative analysis. We firstly pre-train models on UCF101 dataset, then fine-tune the models for video action recognition task.

In Table \ref{tab:clip}, we can see that more input frames bring better performance. As the input length increases, MCN takes additional improvement.

	\begin{table}[htbp]
		\centering
		\resizebox{0.68\linewidth}{!}{%
			\begin{tabular}{|l|c|c|c|}
				\hline 
				\textbf{Methods}  & \textbf{Input Frames} & \textbf{UCF101(\%)}  \\ 
				\hline
				\hline
				Baseline   & 16 & 74.6   \\ %
				\cellcolor{gray!25}MCN  & \cellcolor{gray!25}16 & \cellcolor{gray!25}\textbf{81.3}  \\ %
				\hline
				Baseline  & 32 & 77.3  \\ %
				\cellcolor{gray!25}MCN  & \cellcolor{gray!25}32 &  \cellcolor{gray!25}\textbf{84.8} \\ %
				\hline
				Baseline  & 64 & 80.6  \\ %
				\cellcolor{gray!25}MCN  & \cellcolor{gray!25}64 &  \cellcolor{gray!25}\textbf{86.7} \\ %
				\hline
			\end{tabular}
		}
		\vspace{0.25cm}
		\caption{Results of different input frames for MCN and baseline on video action recognition task.} 
		\label{tab:clip}
	\end{table}

\textbf{Influence of Individual Component.} We also test each component of MCN to figure out their contributions to the final performance. The results of video action recognition on UCF101 are demonstrated in Table \ref{tab:role}. R(2+1)D is selected as backbone. 

As shown in Table \ref{tab:role}. CL represents contrastive loss. BL represents binary loss from proposed meta branch. Combining CL and BL without meta stages takes \textbf{1.9\%} accuracy improvement. By adding meta stages, additional \textbf{5.6\%} improvement is achieved, which proves the efficiency of meta learning. These experiments can demonstrate the effectiveness of proposed MCN method.

	\begin{table}[htbp]
		\centering
		\resizebox{0.7\linewidth}{!}{%
			\begin{tabular}{|c|c|c|c|}
				\hline 
				\textbf{CL} & \textbf{BL} & \textbf{Meta Stages} & \textbf{UCF101(\%)}  \\ 
				\hline
				\hline
				\cmark & & &  77.3 \\ %
				\cmark & \cmark & &  79.2  \\ %
				\cellcolor{gray!25}\cmark & \cellcolor{gray!25}\cmark & \cellcolor{gray!25}\cmark & \cellcolor{gray!25}\textbf{84.8} \\ %
				\hline
			\end{tabular}}
		\vspace{0.3cm}
		\caption{Ablation study for different components of MCN on video recognition task.} 
		\label{tab:role}
	\end{table}
	
\subsection{Evaluation of MCN}
In this section, we compare the performance of our proposed method with other state-of-the-art approaches. We show the evaluations on two downstream tasks including video action recognition and video retrieval. R3D-18 and R(2+1)D are used as backbones for the comparisons.

\textbf{Video Action Recognition.}  Considering that we only use RGB information in our experiments, we didn't include the approaches with multi-modality \cite{korbar2018cooperative, alwassel2019self, patrick2020multi, miech2020end}. CoCLR also \cite{Han20} demonstrates excellent performance by co-training RGB and optical flow samples. In this paper, we only include the RGB-only results of CoCLR for fair comparison. 

We first compare our linear probe evaluation results with other state-of-the-art approaches so that we can verify the transferability of the video representations learned with our approach. Results in Table \ref{tab:ucf101_linear_probe} demonstrate that the proposed MCN method outperforms state-of-the-art approaches on both UCF101 and HMDB51. 

	\begin{table}[htbp]
		\centering
		\resizebox{0.7\linewidth}{!}{%
			\begin{tabular}{|l|c|c|}
				\hline 
				\textbf{Methods}  & \textbf{UCF101(\%)}  & \textbf{HMDB51(\%)}  \\ 
				\hline
				\hline
				CBT \cite{sun2019learning} & 54.0 & 29.5   \\ %
				MemDPC \cite{han2020memory} & 54.1 & 30.5 \\ %
				CoCLR \cite{Han20}  & 70.2  & 39.1 \\ %
				\hline
				\cellcolor{gray!25}Ours (R(2+1)D)  & \cellcolor{gray!25}\textbf{72.4} & \cellcolor{gray!25}\textbf{42.2} \\ %
				\cellcolor{gray!25}Ours (R3D-18)  & \cellcolor{gray!25}\textbf{73.1} & \cellcolor{gray!25}\textbf{42.9} \\ %
				\hline
			\end{tabular}
		}
		\vspace{0.25cm}
		\caption{Linear probe comparisons with state-of-the-art methods on UCF101 and HMDB51 datasets.}
		\label{tab:ucf101_linear_probe}
	\end{table}
	
	\begin{table}[htbp]
		\centering
		\resizebox{\columnwidth}{!}{%
			\begin{tabular}{|l|c|c|c|c|c|c|}
				\hline 
				\textbf{Methods} & \textbf{Backbone} & \textbf{Resolution} & \textbf{UCF101}  & \textbf{HMDB51}  \\ 
				\hline
				\hline
				Jigsaw\cite{noroozi2016unsupervised} &  UCF101   & 225 & 51.5  & 22.5   \\   %
				OPN \cite{lee2017unsupervised}  &  VGG & 227 &  56.3  & 22.1   \\   %
				Mars \cite{wang2019self} & C3D & 112  & 58.8  & 32.6 \\ %
				CMC \cite{tian2019contrastive} & CaffeNet & 128 & 59.1  & 26.7 \\ %
				ST-puzzle \cite{kim2019self} &R3D  &  224 & 65.0  & 31.3 \\ %
				VCP \cite{luo2020video} &  R(2+1)D &112 & 66.3  & 32.2 \\ %
				VCOP \cite{xu2019self} & R(2+1)D & 112 & 72.4  & 30.9 \\ %
				PRP \cite{yao2020video} & R(2+1)D & 112 & 72.1  & 35.0 \\ %
				IIC \cite{tao2020self} & R3D & 112 & 74.4  & 38.3 \\ %
				PP \cite{wang2020self} & R(2+1)D & 112 & 75.9  & 35.9 \\ %
				CoCLR \cite{Han20}  & S3D & 128 & 81.4  & 52.1 \\ %
				\hline
				\cellcolor{gray!25}Ours  & \cellcolor{gray!25}R(2+1)D &  \cellcolor{gray!25}128 &  \cellcolor{gray!25}\textbf{84.8}  & \cellcolor{gray!25}\textbf{54.5} \\ %
				\cellcolor{gray!25}Ours  & \cellcolor{gray!25}R3D & \cellcolor{gray!25}128 &  \cellcolor{gray!25}\textbf{85.4}  & \cellcolor{gray!25}\textbf{54.8} \\ %
				\hline
			\end{tabular}
		}
		\vspace{0.12cm}
		\caption{Comparisons with state-of-the-art methods for video action recognition on UCF101 and HMDB51 datasets (models are pre-trained on UCF101).}
		\label{tab:ucf101}
		\vspace{0.1cm}
	\end{table}

	\begin{table}[htbp]
		\centering
		\resizebox{\columnwidth}{!}{%
			\begin{tabular}{|l|c|c|c|c|}
				\hline 
				\textbf{Methods} & \textbf{Backbone} & \textbf{Resolution} & \textbf{UCF101}  & \textbf{HMDB51}  \\ 
				\hline
				\hline
				3D-RotNet \cite{jing2018self} & R3D &112 & 62.9  & 33.7 \\ %
				ST-Puzzle\cite{kim2019self} & R3D & 224& 63.9 & 33.7 \\ %
				DPC \cite{han2019video} & R2D-3D &128 & 75.7 & 35.7 \\ %
				SpeedNet \cite{benaim2020speednet} & S3D-G & 224& 81.1 & 48.8 \\ %
				PP \cite{wang2020self} & R(2+1)D & 112 & 75.9  & 35.9 \\ %
				CoCLR \cite{Han20}  & S3D & 128 & 87.9  & 54.6 \\ %
				CVRL \cite{qian2020spatiotemporal}  & R3D & 224 &  92.1  & 65.4 \\ %
				\hline
				\cellcolor{gray!25}Ours  & \cellcolor{gray!25}R(2+1)D & \cellcolor{gray!25}128   &  \cellcolor{gray!25}\textbf{89.2} &  \cellcolor{gray!25}\textbf{58.8}\\ %
				\cellcolor{gray!25}Ours  & \cellcolor{gray!25}R3D  & \cellcolor{gray!25}128 &  \cellcolor{gray!25}\textbf{89.7}  & \cellcolor{gray!25}\textbf{59.3} \\ %
				\hline
			\end{tabular}
		}
		\vspace{0.08cm}
		\caption{Comparisons with state-of-the-art methods for video action recognition on UCF101 and HMDB51 datasets (models are pre-trained on Kinetics-400).}
		\label{tab:kinetics-400}
		\vspace{-0.1cm}
	\end{table}
	
We then compare the results of fine-tuning all parameters with other state-of-the-art methods with different pre-training datasets. In specific, we pre-train our models on both UCF101 and Kinetics-400, and then fine-tune the pre-trained models on UCF101 and HMDB51. Table \ref{tab:ucf101} and Table \ref{tab:kinetics-400} show the results respectively. From the tables, we can observe that the results pre-trained on Kinetics-400 are much better than that pre-trained on UCF101. Kinetics contains much more videos than UCF101. The results demonstrates that MCN can better leverage large volume of unlabeled videos. In both tables, our method outperforms or achieves comparable performance with other state-of-the-art self-supervised approaches. In Table \ref{tab:kinetics-400}, CVRL shows better result than ours. This may be due to three reasons: (1) larger input image resolution (224$\times$224) compared with ours (128$\times$128); (2) more powerful and deeper backbone network (R3D-50) than ours (R(2+1)D and R3D-18); (3) more efficient data augmentation approaches. These experimental results can shed a light for combining meta learning with self-supervised learning approaches.

	\begin{table}[htbp]
		\centering
		\resizebox{\columnwidth}{!}{%
			\begin{tabular}{|l|c|c|c|c|c|}
				\hline 
				\textbf{Methods} & \textbf{top1}  & \textbf{top5}  & \textbf{top10} & \textbf{top20} & \textbf{top50} \\ 
				\hline
				\hline
				Jigsaw \cite{noroozi2016unsupervised} & 19.7 & 28.5  & 33.5 & 40.0 & 49.4 \\ %
				OPN\cite{lee2017unsupervised} & 19.9 & 28.7 & 34.0 & 40.6 & 51.6 \\ %
				Büchler \cite{buchler2018improving} & 25.7 & 36.2 & 42.2 & 49.2 & 59.5 \\ %
				VCOP \cite{xu2019self} & 10.7 & 25.9  & 35.4 & 47.3 & 63.9 \\ %
				VCP \cite{luo2020video}  & 19.9 & 33.7 & 42.0 & 50.5 & 64.4 \\ %
				CMC \cite{tian2019contrastive} & 26.4 & 37.7  & 45.1 & 53.2 & 66.3 \\ %
				PP \cite{wang2020self} & 31.9 & 49.7  & 59.2 & 68.9 & 80.2 \\ %
				IIC \cite{tao2020self} & 42.4 & 60.9 & 69.2 & 77.1  & 86.5 \\ %
				CoCLR \cite{Han20} & 53.3 & 69.4 & 76.6 & 82.0  & - \\ %
				\hline
				\cellcolor{gray!25}Ours (R(2+1)D) & \cellcolor{gray!25}52.5  & \cellcolor{gray!25}\textbf{69.5} & \cellcolor{gray!25}\textbf{77.9} & \cellcolor{gray!25}\textbf{83.1} & \cellcolor{gray!25}\textbf{89.3} \\ %
				\cellcolor{gray!25}Ours (R3D) & \cellcolor{gray!25}\textbf{53.8} & \cellcolor{gray!25}\textbf{70.2} & \cellcolor{gray!25}\textbf{78.3}  & \cellcolor{gray!25}\textbf{83.4}  & \cellcolor{gray!25}\textbf{89.7} \\ %
				\hline
			\end{tabular}
		}
		\vspace{0.05cm}
		\caption{Comparisons with state-of-the-art approaches for video retrieval on UCF101 dataset.}
		\label{tab:video_retrieval_ucf101}
		\vspace{-0.2cm}
	\end{table}

	\begin{table}[htbp]
		\centering
		\resizebox{\columnwidth}{!}{%
			\begin{tabular}{|l|c|c|c|c|c|}
				\hline 
				\textbf{Methods} & \textbf{top1}  & \textbf{top5}  & \textbf{top10} & \textbf{top20} & \textbf{top50} \\ 
				\hline
				\hline
				VCOP \cite{xu2019self} & 7.6 & 22.9  & 34.4 & 48.8 & 68.9 \\ %
				VCP \cite{luo2020video}  & 7.6 & 24.4 & 36.3 & 53.6 & 76.4 \\ %
				CMC \cite{tian2019contrastive} & 10.2 & 25.3  & 36.6 & 51.6 & 74.3 \\ %
				PP \cite{wang2020self} & 12.5 & 32.2  & 45.4 & 61.0 & 80.7 \\ %
				IIC \cite{tao2020self} & 19.7 & 42.9 & 57.1 & 70.6  & 85.9 \\ %
				CoCLR \cite{Han20} & 23.2 & 43.2 & 53.5 & 65.5  & - \\ %
				\hline
				\cellcolor{gray!25}Ours (R(2+1)D) & \cellcolor{gray!25}\textbf{23.7} & \cellcolor{gray!25}\textbf{46.5} & \cellcolor{gray!25}\textbf{58.9} & \cellcolor{gray!25}\textbf{72.4} & \cellcolor{gray!25}\textbf{87.3} \\ %
				\cellcolor{gray!25}Ours (R3D) & \cellcolor{gray!25}\textbf{24.1} & \cellcolor{gray!25}\textbf{46.8} & \cellcolor{gray!25}\textbf{59.7} &\cellcolor{gray!25}\textbf{74.2} & \cellcolor{gray!25}\textbf{87.6}\\ %
				\hline
			\end{tabular}
		}
		\vspace{0.07cm}
		\caption{Comparisons with state-of-the-art approaches for video retrieval on HMDB51 dataset.}
		\label{tab:video_retrieval_hmdb51}
	\end{table}

		
\textbf{Video Retrieval.} In addition to video action recognition task, we also evaluate the performance of MCN on video retrieval task, which can better reflect the semantic-level learning capability. Instead of using RGB and residual views, RGB and flow views of original video clips are considered for video retrieval, in which selected flow view is the vertical dimension of optical flow. We extract optical flow of input videos by using un-supervised TV-L1 algorithm \cite{geiger2012we}. Video retrieval task is conducted with extracted features from pre-trained models without extra fine-tuning stages. We take every video from test set to query $k$ nearest videos from the training set based on its extracted features. When the class of retrieval video is the same as that of the qeury video, this retrieval result is considered correct. The top-1, top-5, top-10, top-20, and top-50 retrieval accuracies have been shown in our experiments.

\begin{figure*}
\begin{center}
\includegraphics[scale=0.59]{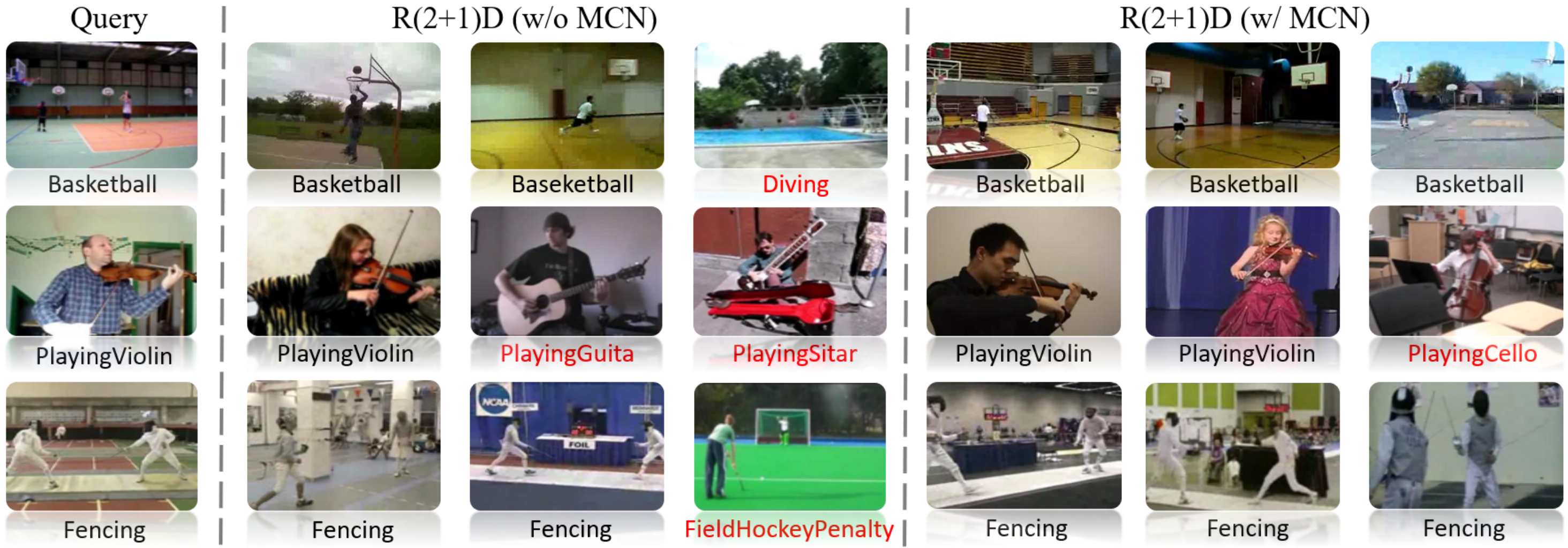}
\end{center}
		\vspace{-0.3cm}
   \caption{\textbf{Video retrieval examples on UCF101.} The first column represents query videos from the test split, and the remaining columns are top-3 results retrieved by the models trained without and with MCN from the training split. The class name of each video is shown in bottom. Red fonts denote wrong video retrieval results.} 
\label{fig:retrieval}
\end{figure*}

As shown in Table \ref{tab:video_retrieval_ucf101}, when compared with other state-of-art methods, our method achieves superior or comparable performance in UCF101 dataset. We observe that the top-1 accuracy of CoCLR is slightly better than our R(2+1)D backbone. Actually, our method is orthogonal to CoCLR. In other words, MCN can take the model trained by CoCLR as baseline to take additional improvement. The results of HMDB51 dataset have been shown in Table \ref{tab:video_retrieval_hmdb51}, which  demonstrate the superior performance of our proposed MCN.

\begin{figure}[H]
\centering
\includegraphics[scale=0.354]{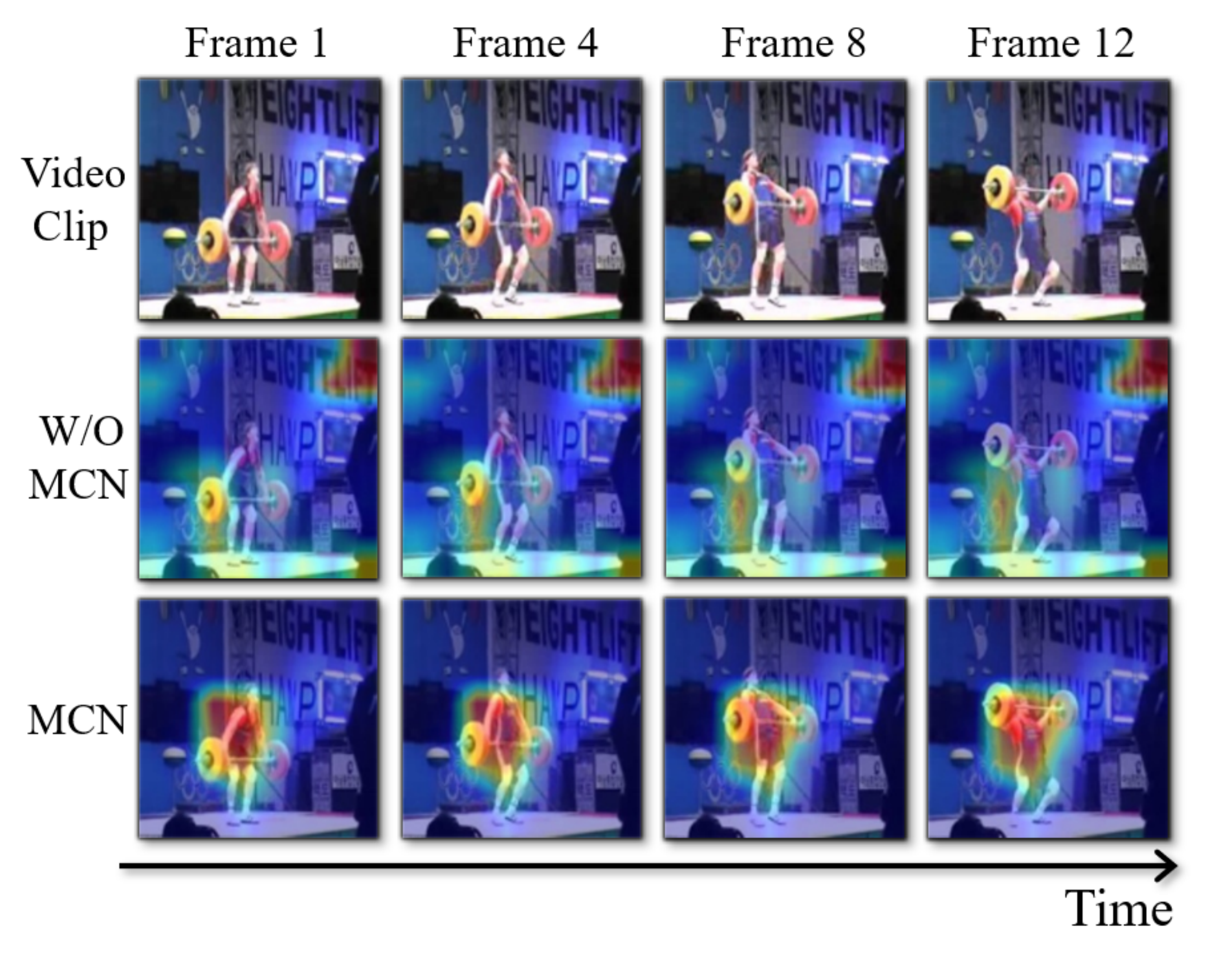}
\caption{\textbf{Activation maps produced from conv5 layer of R(2+1)D backbone.} The maps are generated with 32-frames input and the method in \cite{zagoruyko2016paying} is used. 1$st$, 4$th$, 8$th$ and 12$th$ frames of the video clip are illustrated. The three rows represent original video clip, activation maps produced by models trained without and with MCN respectively. }
\label{fig:vis_sample}
\centering
\end{figure}

\vspace{-0.1cm}
In Figure \ref{fig:retrieval}, retrieval results of models with and without MCN are visualized. R(2+1)D is used as backbone. Video clips from UCF101 test set are used to query 3 nearest videos from UCF101 training set. We can clearly observe that the learned representations with MCN can query videos with the same classes more accurately.

\subsection{Visualization}
In this section, we visualize activation maps of MCN in Figure \ref{fig:vis_sample}, so that we can intuitively understand what has been improved during self-supervised learning process. We use the method in \cite{zagoruyko2016paying} to visualize activation maps from conv5 layer of pre-trained R(2+1)D backbone.

It is interesting to observe that, the model trained without MCN may focus on the irrelevant areas, while MCN can accurately pay attention to the motion areas of video clips. This is essential for action recognition. For example, in the first row of Figure \ref{fig:vis_sample}, we can clearly see that a person is doing a clean and jerk. The learned representations by MCN can focus more on his action areas, such as hands and shoulders.

\section{Conclusion}
In this paper, we propose a novel \textbf{M}eta-\textbf{C}ontrastive \textbf{N}etwork (MCN), which leverages meta-learning to improve the generalization and adaptation ability of contrastive self-supervised video learning on downstream tasks. The proposed MCN contains two branches, \textit{i.e.}, contrastive branch and meta branch, which combine NCE loss and binary classification loss together to enhance the instance discrimination. Meanwhile, we design a two-stage training process based on MAML to improve the learning capability of MCN. Our method outperforms state-of-the-art methods and achieves significant performance boost. To our best knowledge, this is the first time that contrastive self-supervised video learning is combined with meta learning. We also hope our work can inspire more researchers who have interests on this field. 

{\small
\bibliographystyle{ieee_fullname}
\bibliography{egpaper}
}

\end{document}